# Machine learning based radiative parameterization scheme and its performance in operational reforecast experiments

Jing Hao[1,2], Xiao Sa[1,2]*, Li Haoyu[1,2], Xiao Huadong[1,2], Xue Wei[2,3]

[1] State Key Laboratory of Severe Weather Meteorological Science and Technology (LaSW), Beijing, China
[2] CMA Earth System Modeling and Prediction Centre (CEMC), Beijing, China
[3] Tsinghua University, Beijing, China

**Abstract:**
Radiation is typically the most time-consuming physical process in numerical models. One solution is to use machine learning methods to simulate the radiation process to improve computational efficiency. From an operational standpoint, this study investigates critical limitations inherent to hybrid forecasting frameworks that embed deep neural networks into numerical prediction models, with a specific focus on two fundamental bottlenecks: coupling compatibility and long-term integration stability. A residual convolutional neural network is employed to approximate the Rapid Radiative Transfer Model for General Circulation Models (RRTMG) within the global operational system of China Meteorological Administration. We adopted an offline training and online coupling approach. First, a comprehensive dataset is generated through model simulations, encompassing all atmospheric columns both with and without cloud cover. To ensure the stability of the hybrid model, the dataset is enhanced via experience replay, and additional output constraints based on physical significance are imposed. Meanwhile, a LibTorch-based coupling method is utilized, which is more suitable for real-time operational computations. The hybrid model is capable of performing ten-day integrated forecasts as required. A two-month operational reforecast experiment demonstrates that the machine learning emulator achieves accuracy comparable to that of the traditional physical scheme, while accelerating the computation speed by approximately eightfold.

## 1. Introduction

In recent years, machine learning methods have exhibited exceptional performance in addressing problems characterized by high dimensionality and large datasets, while also bringing new vitality and challenges to the field of numerical weather prediction. Currently, meteorological operational and research institutions worldwide have initiated relevant research programs to explore the application of artificial intelligence (AI) technology in weather and climate prediction (ECMWF, 2025). Existing studies have integrated AI methods across the entire workflow, including observation operators (Liang et al., 2023), satellite and radar data processing (Ukamaka et al., 2024), physical parameterization (Chen et al., 2023), data assimilation (Arcucci et al., 2021), ensemble forecasting (Zhong et al., 2024), and post-processing (McGovern et al., 2017).

Furthermore, a growing body of research has applied AI technology to improve weather forecasts in fields like precipitation (Chen et al., 2023), tropical cyclones (Wu et al., 2022), and typhoons (Niu et al., 2024). These methods have demonstrated significant potential to enhance prediction accuracy and boost computational efficiency.

Machine learning technology performs well in identifying and learning complex nonlinear ralationships within datasets (Schmidhuber, 2015). Theoretically, deep neural networks possess the capability to approximate any function (Cybenko, 1989). It implies that they can also model the equations or functions describing atmospheric motion. With the increasing adoption of machine learning in atmospheric science, researchers have explored the development of new parameterization schemes. These schemes leverage machine learning to balance computational efficiency and accuracy, enabling them to not only accurately identify and extract complex nonlinear relationships inherent in physical processes but also address the bottleneck of excessively high computational costs faced by numerical model development (Krasnopolsky, 2020). The use of neural network models to replace or optimize physical process parameterization schemes falls under the category of general form regression tasks. Specifically, neural networks are trained to learn the ralationships between sets of input and output vector pairs. Once trained, they are coupled with numerical models to achieve model replacement or optimization. Such studies can be divided into two categories. The first aims to simulate existing physical parameterization schemes, leveraging the high-efficiency fitting capability of neural networks to reduce computational costs. The second improves physical parameterization schemes by incorporating additional information. For instance, using observation data from satellites, radars and other platforms, or training on higher-resolution models to develop new models to enhance computational accuracy. This hybrid modeling approach (Reichstein et al., 2019), which integrating numerical methods with data-driven techniques, exemplifies the trend and impact of interdisciplinary collaboration across artificial intelligence, high-performance computing, and numerical weather prediction.

Over two decades ago, studies begun to use trained neural networks for simulating radiative parameterization schemes. For example, Chevallier et al. (2000) employed a multi-layer perceptron to replicate the radiation scheme of the ECMWF Integrated Forecast System (IFS) model, while Krasnopolsky et al. (2005) developed a neural network model for application in the National Center for Atmospheric Research (NCAR) CAM2 model. These efforts yielded results comparable to the original radiative transfer processes and increased computational speed by an order of magnitude. However, the neural network structures used at that time were simplistic, preventing full simulation of the entire radiation process. In recent years, advances in deep learning technology have enabled the successful development of deep learning based parameterization schemes for physical processes including radiation (Song and Roh, 2021; Ukkonen, 2022), non-orographic gravity wave drag (Chantry et al., 2021), cloud microphysics (Seifert and Rasp, 2020; Gao et al., 2024), convection (Han et al., 2020; Chen et al., 2023), and planetary boundary layer (Wang et al., 2019) using state-of-the-art techniques. Song et al. (2021) developed two compound radiative parameterizations to reduce uncertainty arising from the extrapolation of extreme data beyond training sets. The first leverages an additional neural network to predict heating rate errors for all input variables, while the second uses cloud fraction to estimate the uncertainty of the neural network emulation. Both approaches significantly reduced the forecast error of the hybrid model. Alves et al. (2023) incorporated 3D cloud radiative effects into ecRad (Hogan and Bozzo, 2016, a library used by ECMWF to compute vertical profiles of solar and near-infrared fluxes and heating rates) by testing various model architectures. Some studies have trained and constructed deep learning models based on the results of higher-precision and high-resolution numerical simulations to solve sub-grid physical processes (Gentine et al., 2018; Rasp et al., 2018; Hu et al., 2025). These models not only accurately simulate subgrid processes such as cloud formation and convective motion but also achieve far greater computational efficiency than traditional numerical simulations. This confirms that deep learning can replicate key advantages of cloud-resolving models at lower computational costs, highlighting the promising application of artificial intelligence in substantially enhancing the overall performance of numerical models (Chantry et al., 2021).

On the other hand, deep learning based physical parameterization schemes still need to address critical issues related to stability and physical interpretability. The concept of coupled operation refers to a process where the deep learning parameterization scheme takes inputs from the numerical model to generate predictions, which are then fed as inputs into other model modules. This process iterates continuously to enable time-integrated simulations. Stability during coupled operation is not only a prerequisite for long-term integration but also a core factor for promoting the practical application of hybrid models in operational settings. Numerical instability in this process may stem from non-physical correlations (Brenowitz and Bretherton, 2019), inadequate enforcement of physical constraints (Yuval et al., 2021), and suboptimal selection of model network structures, hyperparameters, and optimizers. As network depth increases, the complexity and abstraction of deep learning models rise sharply. Currently, there is no comprehensive theory to systematically deduce their internal decision-making logic (Hernandez-Orallo, 2021), leading to a lack of transparency and interpretability in these models. Such insufficient interpretability undermines their credibility in numerical forecasting applications. Thus, the interpretability of deep learning models is critical to ensuring their predictions align with physical laws (Toms et al., 2020).

For the practical operational implementation of hybrid models, two critical challenges must be addressed, coupling programming issues and model integration interruption. This study focuses on developing feasible solutions to these two key problems from an operational perspective. We adopts data-driven approaches to model the radiation processes within the Global Forecasting System of China Meteorological Administration (CMA-GFS). Guided by the operational mechanism of the radiation process, we first screened the input and output physical quantities for the deep learning model and constructed a well-designed dataset. This step aims to enable the deep learning model to effectively capture and learn the inherent nonlinear characteristics of radiative processes. Subsequently, we analyzed and summarized the key data features of the dataset, which provides critical support for developing effective deep learning parameterization schemes and facilitating physical interpretation of the model. In accordance with the characteristics and mechanisms of coupled operation, a deep learning parameterization scheme was designed and integrated into the host model. We developed a Libtorch-based library for real-time operations, which compiles the machine learning emulator into a callable shared library. This library is then linked in the compilation and linking phase to generate the executable program. To enhance the stability of the hybrid model, we leverage prior knowledge to reduce the randomness of the machine learning model, and adopt strategies such as constraint imposition and empirical feedback to ensure the stable operation of the hybrid model. This study also presents the performance of the online hybrid model in real operational scenarios.

**2. RRTMG-original radiation scheme in CMA-GFS**

Radiative transfer refers to the transport of energy via electromagnetic waves through a gas medium. The Line-by-Line Radiative Transfer Model (LBLRTM, Clough et al., 1992) serves as the foundational code for the development of all radiation related models. LBLRTM calculates the absorption and emission of radiation by gaseous molecules in the atmosphere, while accounting for the effects of thousands of individual absorption lines. Climate and weather prediction models require fast calculations of radiative flux and heating rates to achieve accurate simulations (Price et al., 2014). The Rapid Radiative Transfer Model (RRTM) has been developed for both the longwave and shortwave regions as reference broadband radiative transfer models that closely reproduce line-by-line results (Mlawer et al., 1997). Absorption coefficients for the primary and minor molecular species required for the correlated k distribution method used by RRTM have been obtained from LBLRTM. Molecular absorbers included in RRTM are water vapor, carbon dioxide, ozone, methane, nitrous oxide, oxygen, nitrogen and the halocarbons in the longwave and water vapor, carbon dioxide, ozone, methane and oxygen in the shortwave.

RRTM retains the highest accuracy relative to line-by-line results for single-column calculations. It has since been modified to develop RRTMG (Iacono et al., 2008), a version optimized to deliver higher computational efficiency with minimal accuracy loss, specifically for general circulation model applications. While RRTMG shares the same fundamental physics and absorption coefficients as RRTM, it incorporates several modifications to improve computational efficiency, and to represent subgrid-scale cloud variability. In particular, the total number of quadrature points (g points) used to calculate radiances in the longwave has been reduced from the standard 256 in RRTM_LW (with 16 g points in each of the 16 spectral bands) to 140 in RRTMG_LW (with the number of g points in each spectral band varying from 2 to 16 depending on the absorption in each band). In the shortwave, the number of g points has been reduced from the standard 224 in RRTM_SW (with 16 in each of the 14 spectral bands) to a total of 112 in RRTMG_SW. RRTM computes multiple scattering with the discrete ordinates algorithm, while RRTMG uses a two-stream algorithm (Oreopoulos and Barker, 1999). Subgrid-scale cloud variability is not considered in RRTM. To improve the realism of cloud-radiation interactions, RRTMG incorporates the Monte-Carlo independent column approximation (McICA) method (Barker et al., 2007).

RRTMG continues to be extensively validated with measured atmospheric spectra from the sub-millimeter to the ultraviolet. Its applications include the IFS system of the ECMWF (Morcrette et al., 2008); the Community Earth System Model and the WRF regional forecast model at NCAR; the CMA-GFS model at China Meteorological Administration Earth System

Modeling and Prediction Centre. Although RRTMG is a fast algorithm for representing atmospheric radiative transfer, it remains the most computationally expensive component among the various physical process parameterizations. If radiative transfer impacts were computed for every grid point at all time steps, the associated computational cost would typically be equivalent to that of the model's remaining dynamical and other physical parameterizations combined. To reduce computational time, the radiation scheme is usually invoked only at large temporal intervals and on a coarsened grid. Subsequently, the radiative effects are interpolated to the finer model grid using a cubic interpolation scheme (Morcrette et al., 2008). Specifically, in the CMA-GFS model, RRTMG is called once per hour and computed on a reduced grid aligned with latitude parallels. Additionally, a load balancing strategy is implemented, as shortwave radiation calculations are only required for half of the globe. Nevertheless, RRTMG still accounts for the largest share of computational time among all physical processes in the model. To address this remaining inefficiency, this study aims to further improve computational efficiency by simulating radiative processes using deep neural networks.

### 3. Dataset

The selection and construction of a training dataset is a critical step prior deep learning modeling, as it directly determines whether the deep learning model can effectively capture and learn the nonlinear characteristics inherent in physical processes. Following the data interaction mechanism between RRTMG and the host model, the input and output variables generated during the operation of the RRTMG scheme are adopted as the training and testing data for the deep learning based parameterization scheme.

For a given atmospheric column, RRTMG takes atmospheric state variables provided by the CMA-GFS model as inputs and computes radiative fluxes and heating rates as outputs. Table 1 and Table 2 present the basic information (e.g., names and dimensions) of the input and output physical variables for longwave and shortwave radiation within a single vertical grid column, respectively. The input variables include pressure, temperature, gas mixing ratios, and cloud-related parameters. Variables with vertical height layers are vectors of length 88 or 89, corresponding to the number of model layers. Cloud-related parameters are associated with different absorption bands and have one additional dimension compared to conventional one-dimensional vectors. In contrast, physical variables without height information are scalars. The output variables comprise all-sky upward/downward radiative fluxes, all-sky heating rates, as well as clear-sky upward/downward radiative fluxes and clear-sky heating rates. To reduce model complexity and account for the influence of different variables on the output radiative fluxes and heating rates, the variables highlighted in red in Tables 1 and Table 2 were selected as the inputs and outputs for the neural network.

For shortwave radiation, an input-output dataset covering the entire year of 2022 was constructed. It has a spatial resolution of 25 km, an output interval of 25 hours, and was generated by consecutively running 5-day forecasts with sequentially incremented dates. The total size of this dataset is approximately 60 TB, designed to fully capture the diverse states of shortwave radiation in real numerical simulations. During model training, the dataset was randomly sampled to form a 4.9 TB training set and a roughly 500 GB validation set. An equivalent input-output dataset was also constructed for longwave radiation with identical spatial resolution (25 km) and output interval (25 hours). It was generated using the same method by consecutively running 5-day forecasts with sequentially incremented dates to cover the entire year of 2022, resulting in a total size of around 76 TB. For model training, this dataset was randomly sampled to create a 4.2 TB training set and an approximately 450 GB validation set.

Radiation schemes are typically one-dimensional, meaning that each atmospheric column is processed independently. Each sample uses a single vertical grid column as the unit and includes information on various types of input and output physical variables. These samples cover a range of physical variables with distinct properties, which introduces heterogeneity into the dataset. When working with heterogeneous data, the first step is to normalize the data. Normalizing input data helps the machine learning model more effectively extract meaningful features, while normalizing output data prevents the optimizer from over-optimizing physical variables with large numerical scales during training. In this study, standard normalization was applied to the original radiation data to reduce differences in the orders of magnitude across different physical variables.

Table 1. Statistics of longwave radiation variables, variables highlighted in red serve as inputs and outputs of the neural network.

| Inputs | | | |
|---|---|---|---|
| **Variables** | **Meaning and dimensions** | **Variables** | **Meaning and dimensions** |
| ncol | Number of horizontal column | cfc11vmr | CFC11 volume mixing ratio (ncol, naly) |
| nlay | Number of model layers | cfc12vmr | CFC12 volume mixing ratio (ncol, nlay) |
| icld | Cloud overlap method | cfc22vmr | CFC22 volume mixing ratio (ncol, nlay) |
| idrv | Flag for calculation of dFdT | ccl4vmr | CCL4 volume mixing ratio (ncol, nlay) |
| play | Layer pressures (hPa, mb) (ncol, nlay) | emis | Surface emissivity (ncol, nbndlw) |
| plev | Interface pressures (hPa, mb) (ncol, nlay+1) | inflglw | Flag for cloud optical properties |
| tlay | Layer temperatures (K) (ncol, nlay) | iceflglw | Flag for ice particle specification |
| tlev | Interface temperatures (K) (ncol, nlay+1) | liqflglw | Flag for liquid droplet specification |
| tsfc | Surface temperatures (K) (ncol) | tauaer | Aerosol optical depth at mid-point of LW spectral bands (ncol, nlay, nbndlw) |
| h2ovmr | H2O volume mixing ratio (ncol, nlay) | cldfmcl | Cloud fraction (ngptlw, ncol, nlay) |
| o3vmr | O3 volume mixing ratio (ncol, nlay) | taucmcl | In-cloud optical depth (ngptlw, ncol, nlay) |
| co2vmr | CO2 volume mixing ratio (ncol, nlay) | ciwpmcl | In-cloud ice water path (ngptlw, ncol, nlay) |
| ch4vmr | Mathane volume mixing ratio (ncol, nlay) | clwpmcl | In-cloud liquid water path (ngptlw, ncol, nlay) |
| n2ovmr | Nitrous oxide volume mixing ratio (ncol, nlay) | reicmcl | Cloud ice particle effective size (ncol, nlay) |
| o2vmr | Oxygen volume mixing ratio (ncol, nlay) | relqmcl | Cloud water drop effective radius (ncol, nlay) |

| Outputs | |
|---|---|
| **Variables** | **Meaning and dimensions** |
| uflx | Total sky longwave upward flux (W/m2) (ncol, nlay+1) |
| dflx | Total sky longwave downward flux (W/m2) (ncol, nlay+1) |
| hr | Total sky longwave radiative heating rate (K/d) (ncol, nlay) |
| uflxc | Clear sky longwave upward flux (W/m2) (ncol, nlay+1) |
| dflxc | Clear sky longwave downward flux (W/m2) (ncol, nlay+1) |
| hrc | Clear sky longwave radiative heating rate (K/d) (ncol, nlay) |
| lw_down | Downward longwave surface flux |
| duflx_dt | Change in upward longwave flux (w/m2/K) (ncol, nlay) |
| duflxc_dt | Change in clear sky upward longwave flux (w/m2/K) (ncol, nlay) |

Table 2. Statistics of shortwave radiation variables, variables highlighted in red serve as inputs and outputs of the neural network.

| Inputs | | | |
|---|---|---|---|
| **Variables** | **Meaning and dimensions** | **Variables** | **Meaning and dimensions** |
| ncol | Number of horizontal column | adjes | Flux adjustment for earth/sun distance |
| nlay | Number of model layers | coszen | Cosine of solar zenith angle (ncol) |
| icld | Cloud overlap method | scon | Solar constant |
| play | Layer pressures (hPa, mb) (ncol, nlay) | inflgsw | Flag for cloud optical properties |
| plev | Interface pressures (hPa, mb) (ncol, nlay+1) | iceflgsw | Flag for ice particle specification |
| tlay | Layer temperatures (K) (ncol, nlay) | liqflgsw | Flag for liquid droplet specification |
| tlev | Interface temperatures (K) (ncol, nlay+1) | cldfmcl | Cloud fraction (ngptsw, ncol, nlay) |
| tsfc | Surface temperatures (K) (ncol) | taucmcl | In-cloud optical depth (ngptsw, ncol, nlay) |
| h2ovmr | H2O volume mixing ratio (ncol, nlay) | ssacmcl | In-cloud single scattering albedo (ngptsw, ncol, nlay) |
| o3vmr | O3 volume mixing ratio (ncol, nlay) | asmcmcl | In-cloud asymmetry parameter (ngptsw, ncol, nlay) |
| co2vmr | CO2 volume mixing ratio (ncol, nlay) | fsfcmcl | In-cloud forward scattering fraction (ngptsw, ncol, nlay) |
| ch4vmr | Mathane volume mixing ratio (ncol, nlay) | ciwpmcl | In-cloud ice water path (g/m2) (ngptsw, ncol, nlay) |
| n2ovmr | Nitrous oxide volume mixing ratio (ncol, nlay) | clwpmcl | In-cloud liquid water path (g/m2) (ngptsw, ncol, nlay) |
| o2vmr | Oxygen volume mixing ratio (ncol, nlay) | reicmcl | Cloud ice effective radius (microns) (ncol, nlay) |
| asdir | UV/vis surface albedo direct rad (ncol) | relqmcl | Cloud water drop effective radius (ncol, nlay) |
| aldir | Near-IR surface albedo direct rad (ncol) | tauaer | Aerosol optical depth (ncol, nlay, nbndsw) |
| asdif | UV/vis surface albedo: diffuse rad (ncol) | ssaaer | Aerosol single scattering albedo (ncol, nlay, nbndsw) |
| aldif | Near-IR surface albedo: diffuse rad (ncol) | asmaer | Aerosol asymmetry parameter (ncol, nlay, nbndsw) |
| dyofyr | Day of the year | ecaer | Aerosol optical depth at 0.55 micron (ncol, nlay, naerec) |

| Outputs | |
| --- | --- |
| **Variables** | **Meaning and dimensions** |
| swuflx | Total sky shortwave upward flux (W/m2) (ncol, nlay+1) |
| swdflx | Total sky shortwave downward flux (W/m2) (ncol, nlay+1) |
| swhr | Total sky shortwave radiative heating rate (K/d) (ncol, nlay) |
| swuflxc | Clear sky shortwave upward flux (W/m2) (ncol, nlay+1) |
| swdflxc | Clear sky shortwave downward flux (W/m2) (ncol, nlay+1) |
| swhrc | Clear sky shortwave radiative heating rate (K/d) (ncol, nlay) |
| sw_down_vis_dir | Downward shortwave surface flux, direct, UV/vis |
| sw_down_vis_dif | Downward shortwave surface flux, diffuse, UV/vis |
| sw_down_nir_dir | Downward shortwave surface flux, direct, near-IR |
| sw_down_nir_dif | Downward shortwave surface flux, diffuse, near-IR |

**4. Machine learning model and training**

The machine learning based physical parameterization scheme processes data using independent vertical grid columns as the unit. Consequently, both input and output data take the form of one-dimensional vectors, each composed of multiple physical variables. In this study, we use a convolutional neural network (CNN) as the base model and incorporate residual modules (He et al., 2016). It allows the deep network to effectively learn complex nonlinear information embedded in the data.

In this study, input and output variables within a single vertical grid column were stacked separately. By merging the different absorption bands of cloud-related parameters, a 22-channel one-dimensional vector was used as input and a 6-channel one-dimensional vector was used as output for shortwave radiation. For longwave radiation, the input was a 20-channel one-dimensional vector, and the output was a 6-channel one-dimensional vector. Since some input variables are defined on full model layers and others on half layers, the data were padded to ensure a uniform 89-layer structure. Based on the input and output dimensions, a deep learning parameterization scheme was constructed using RseCNN to predict the radiative fluxes and heating rates within each vertical grid column. This is consistent with the operational principle of the RRTMG scheme. The model incorporates three residual modules, each containing two convolutional layers. Additional fully connected layers were added to capture more global information of the single column, as illustrated in Figure 1. Other neural networks were also tested, including fully connected neural networks and convolutional block attention networks (CBAMs, Woo et al., 2018). However, when the number of parameters was comparable to that of ResCNN, the fully connected model exhibited relatively poor representational capability. Meanwhile, the CBAM model did not improve the prediction accuracy despite having more parameters.

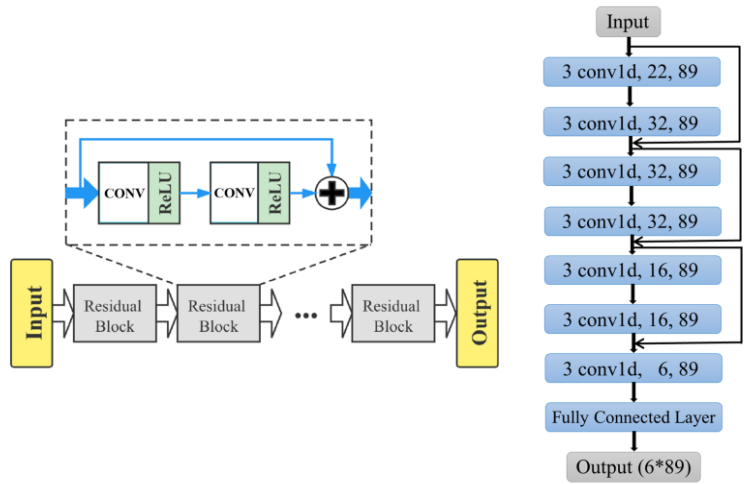

Figure 1. schematic diagram of the longwave radiation machine learning model.

Model training was conducted using P100 GPUs and based on the PyTorch framework. Given the 'small model, large dataset' configuration adopted for this task, the training process was implemented via Distributed Data Parallelism, which is a built-in distributed data-parallel training module in PyTorch. Taking shortwave radiation as an example, for a 4.9TB training set and a 500GB validation set, training was performed using 8 P100 GPUs, lasting 68 hours and completing a total 80 epochs. The ReLU function was used as the activation function, the initial learning rate was set to 0.005, with cosine annealing adopted as the learning rate adjustment strategy, and the SGD optimizer was employed.

First we conducted offline tests on the radiation machine learning (ML) emulator. On the test set, the discrepancies between the model variables predicted by the neural network and the reference values (from the model output) were relatively small. As shown in Figures 2 and Figure 3, the left panel compares the six variables predicted by the neural network with the reference values output by RRTMG, while the right panel presents the mean absolute error (MAE) of these six variables and their biases across different percentiles. Although biases are relatively larger at the bottom and top layers of the model, the fitting accuracy exceeds 99% in most cases. This indicates that the trained neural network can effectively approximate the radiation process.

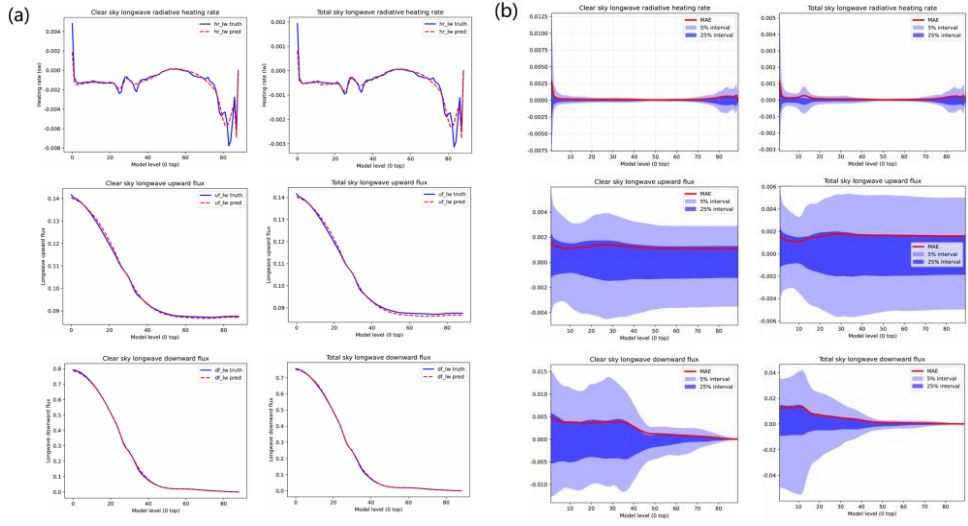

Figure 2. Offline test results of the longwave radiation, (a) comparisons between the predicted variables of the neural network and the reference values, (b) MAE and 5th/25th percentile confidence intervals.

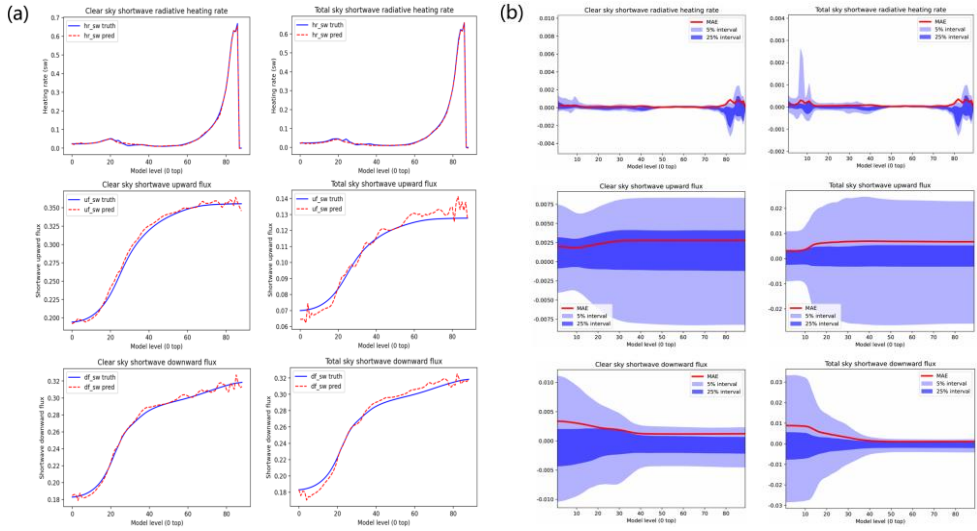

Figure 3. Offline test results of the shortwave radiation, (a) comparisons between the predicted variables of the neural network and the reference values, (b) MAE and 5th/25th percentile confidence intervals.

## 5. Coupling the ML emulator and CMA-GFS

Deep learning parameterization schemes are mainly trained using Python-based machine learning frameworks. Python is a dynamically typed language, which means deep learning models cannot be directly invoked as functions by numerical models developed in Fortran, a statically typed language.

To couple the ML emulator with traditional numerical models, one approach is to hardcode deep learning models into Fortran code. However, this method has low development efficiency.

Any modification to the deep learning model requires corresponding changes to the associated code. To address this, researchers have developed an interface called the fortran-keras deep learning bridge (Otte et al., 2020). This interface enables Fortran programs to import pre-training deep learning parameters (i.e., weights and biases) from external sources. These parameters are stored in TXT files and invoked when the deep learning model performs inference. Notably, the interface can be deployed in numerical models with relative flexibility, as it does not require modifying the deep learning model's network structure. Users only need to update the parameters in the TXT file to replace the model. Nevertheless, this interface primarily supports models based on fully connected layers and cannot accommodate many state-of-the-art network architectures, such as diffusion models or Transformer models.

Another widely used coupling method splits the Fortran-based host model and Python-based deep learning components into two separate process sets. It enables data transmission via Inter-Process Communication (IPC) while achieving synchronous control during coupled operations (Wang et al., 2022). During coupled operations, the IPC data interface first receives input variables from the host model and performs a gather operation on the raw data to enhance data transmission efficiency. Next, it conducts floating-point endian conversion and sends the data to the ML emulator using IPC. Notably, the IPC data interface supports multiple communication modes, including Message Passing Interface (MPI), First-In-First-Out (FIFO) pipes, and high-speed data buffers. After the ML emulator completes its computations, it first writes the results back to the IPC data interface. The interface then performs gather operations and floating-point endian conversion before returning the results to the host model. Throughout the entire data coupling process, a synchronization controller manages the operational synchronization between the host model and the ML emulator. While this coupling method is ideal for online learning and the initial debugging of model coupling, communication between Fortran and Python processes introduces extra computational latency.

The Fortran Torch Adapter (FTA) is a versatile tool built on LibTorch (Mu et al., 2023), with a design philosophy similar to that of the ECMWF coupling library Infero (https://infero.readthedocs.io/en/latest/index.html). This tool supports converting deep learning models under the PyTorch framework into a static binary format and provides a C/C++ interface. Subsequently, Fortran-C/C++ mixed programming techniques are used to deploy the ML emulator in numerical models.

Integrating ML emulator into the host model involves frequent calls to identical small-scale networks for fast inference and frequent communication. This scenario is highly suited to the LibTorch-based coupling method for real-time operation. We developed an analogous LibTorch-based compilation library tool. Acting as an intermediate layer, the compilation library tool offers Fortran users a simple, transparent interface for operating PyTorch models, while hiding the underlying complexity of connecting to C++ interfaces via language interoperability. Users only need to invoke these interfaces like regular Fortran library functions. The compilation library tool automatically handles all trivial inconsistencies between Fortran and C++, such as data types and array memory models. This tool also supports customization based on user-specific needs, with customizable parameters including system environment configurations (e.g., the compiler used and PyTorch version) and usage options (e.g., the dimensions and types of input/output data for the ML emulator, whether to utilize GPU resources, and the ML emulator's storage location). This coupling method provides high flexibility. However, it still faces certain code compatibility issues in mixed programming. Additionally, since LibTorch is based on C/C++ programming standards, debugging and stability testing during coupled operations remain less flexible and efficient.

Subsequently, by writing code at the locations where network activation is required, we can enable calls to the PyTorch network trained offline. In this setup, the numerical model and the ML emulator perform computations within the same process, eliminating the need for additional communication. The mechanism of the compilation library tools is illustrated in Figure 4.

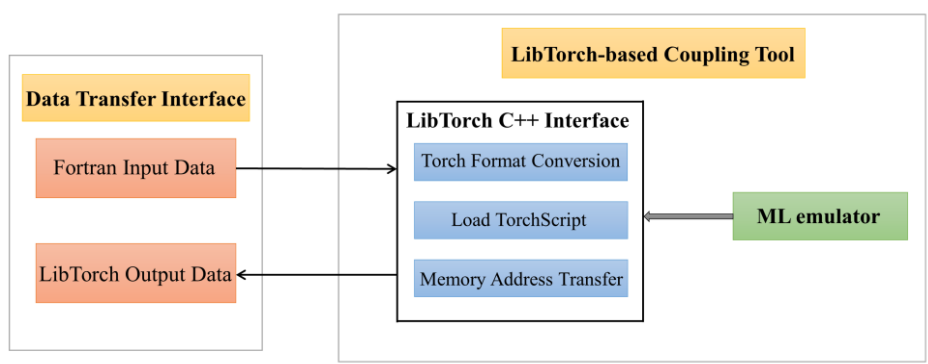

Figure 4. The mechanism of the LibTorch-based coupling tool.

## 6. Enhance the stability of the hybrid model

The radiation ML emulator is coupled with the CMA-GFS model via a compiled library tool, resulting in a hybrid model. Once the network is sufficiently trained, the deep learning model can approximate the accurate solutions of radiative processes generated by RRTMG. Due to the application of standard normalization, the output values of the deep learning model are constrained within the interval 0 and 1. These values are then multiplied by a normalization coefficient to generate the final output, which is transmitted to the host model. For radiation processes, although the training data were generated at a 25 km resolution, the hybrid model can still operate at a 12.5 km resolution and support 10-day integral forecasts required for operational purposes.

For the hybrid model, when both the ML emulator and the host model run on CPU nodes, each process handles an equal number of samples. As shown in Table 3, the computational efficiency of the ML emulator is improved by nearly eightfold on average compared with the original RRTMG. Furthermore, we compare the time consumption of the ML emulator and the original RRTMG module on different parallel computing scales. It can be seen that RRTMG has excellent scalability. The ML emulator also exhibits good scalability, though slightly inferior to that of the physical model. When the ML emulator is executed on a GPU node, its inference speed is inherently enhanced. However, the data transmission latency must be considered.

Table 3. Comparison of computation time: ML emulator in the hybrid model vs. RRTMG in the original model.

| Number of CPU cores | 1920 | 3840 | 7680 | 11520 | 15360 |
|---|---|---|---|---|---|
| Radiation ML emulator | 0.204s | 0.093s | 0.053s | 0.037s | 0.028s |
| RRTMG | 1.441s | 0.715s | 0.342s | 0.228s | 0.177s |

We conducted 50 experiments using the initial version of the hybrid model, with ten-day forecasts initiated at randomly selected times in 2023. Among these experiments, 18% experienced interruptions. When such interruptions occurred, we traced and identified the coupling breakpoints between the ML emulator and the host model. Most interruptions were found to originate from a critical issue: the uncertainty caused by extrapolating extreme data beyond the training set poses a major challenge to the emulator. This uncertainty may lead to the breakdown of the entire model during long-term simulations, particularly in extreme weather conditions or edge scenarios such as typhoons and low-temperature environments in high-latitude regions. The long-term stable coupled operation of the deep learning model and the host model is a prerequisite for transitioning deep learning parameterization schemes form research to practical operational applications.

To address the instability issue, we considered the following solutions. First, leverage prior knowledge and the physical process logic of numerical models to reduce the blindness of machine learning model. For longwave radiation, the heating rate predicted by the ML emulator shows a relatively large error. The required heating rate can be calculated using the following relationship:

$$Heating\_rate[i] = -\frac{g}{c_p} \cdot \frac{flux\_dn[i] + flux\_up[i] - flux\_dn[i+1] - flux\_up[i+1]}{hl\_pressure[i] - hl\_pressure[i+1]},$$

where Heating_rate[i] denotes the atmospheric heating rate at layer i, g represents gravitational acceleration, $c_p$ stands for the specific heat capacity of air at constant pressure, flux_dn[i] is the downward radiative flux at layer i, flux_up[i] is the upward radiative flux at layer i, and hl_pressure[i] refers to the atmospheric pressure at layer i. Second, constrain the output range of the ML emulator by setting reasonable upper and lower bounds for output variables based on their physical significance. This prevents the ML model from generating unphysical predictions caused by input noise or extreme values. Third, train the model using a diverse dataset that covers as many extreme weather events and edge scenarios as possible. This avoids significant deviations of the ML emulator in unseen scenarios. Additionally, random noise is added to the training data to enhance the model's robustness against input perturbations. Testing results indicated that the third solution played the most significant role in maintaining the integration stability of the hybrid model. Specifically, we collected interruption samples from the previous experiments to form a new optimization dataset. By combing this new dataset with the original one to create an updated training dataset, all 50 selected experiments successfully completed the ten-day forecasts without interruptions. This was achieved without modifying the structure of the machine learning models.

**7. Performance in the operational reforecast experiments**

The operational version of the CMA-GFS model runs at a horizontal resolution of 12.5 km. We conducted a two-month operational reforecast experiment on the hybrid model, covering January 2024 and July 2024. The results were compared against those from the operational system. First, we compared the forecast performance of synoptic fields, as shown in Figure 5. The hybrid model and the operational model showed comparable performance in synoptic field forecasting, with a slight reduction in predictable days observed over the Northern Hemisphere in January.

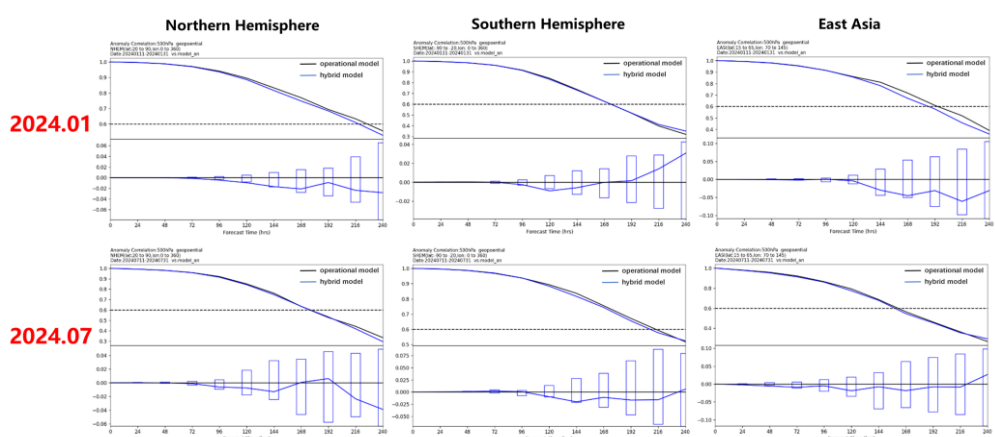

Figure 5. Synoptic fields comparisons between the hybrid and operational models in January and July.

Next, we verify the precipitation forecasts, with results presented in Figure 6 to Figure 9. For the Threat Score (TS), the hybrid model outperformed the operational model in several scenarios. In January, it showed advantages in forecasting light and moderate rain within the first 3 days of lead time. In July, it excelled at predicting heavy and torrential rain within the first 1.5 days. For other lead times and rainfall grades, the hybrid model's precipitation forecasts were either comparable to or less accurate than those of the operational model. Regarding precipitation forecast bias, in January, it exhibited lower bias than the operational model across all rainfall grades within the first 2 days of lead time. However, this advantage reversed beyond the 2-day mark with its bias becoming higher than that of the operational model. In July, the hybrid model showed a more grade-specific pattern. It had higher bias for light and moderate rain, yet lower bias for heavy and torrential rain regardless of lead time. For accumulated precipitation assessment, the hybrid model barely changed the forecast of daily accumulated precipitation below 10 mm compared with the operational model. For daily accumulated precipitation above 10 mm, its precipitation distribution was more consistent with observations, with a slight increase in the forecast of precipitation extremes.

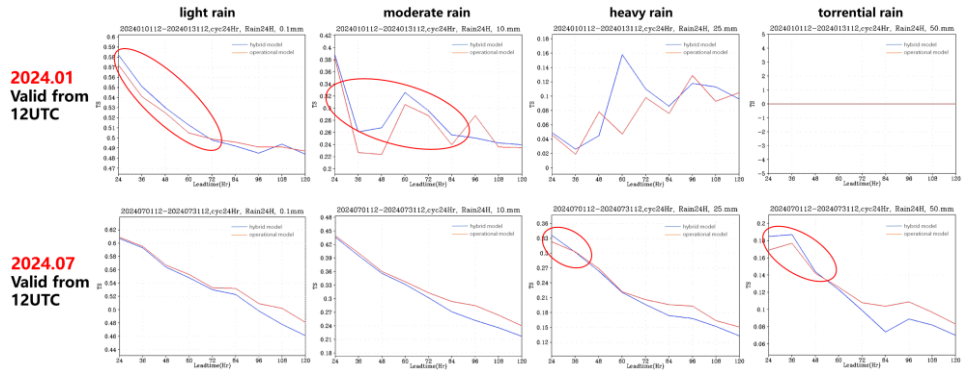

Figure 6. Comparisons of Threat Scores for precipitation forecasts: hybrid model vs. operational model.

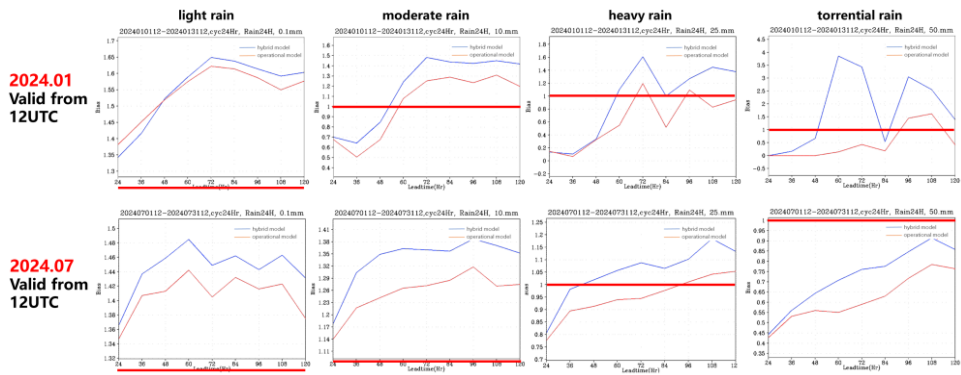

Figure 7. Comparisons of precipitation forecast biases: hybrid model vs. operational model.

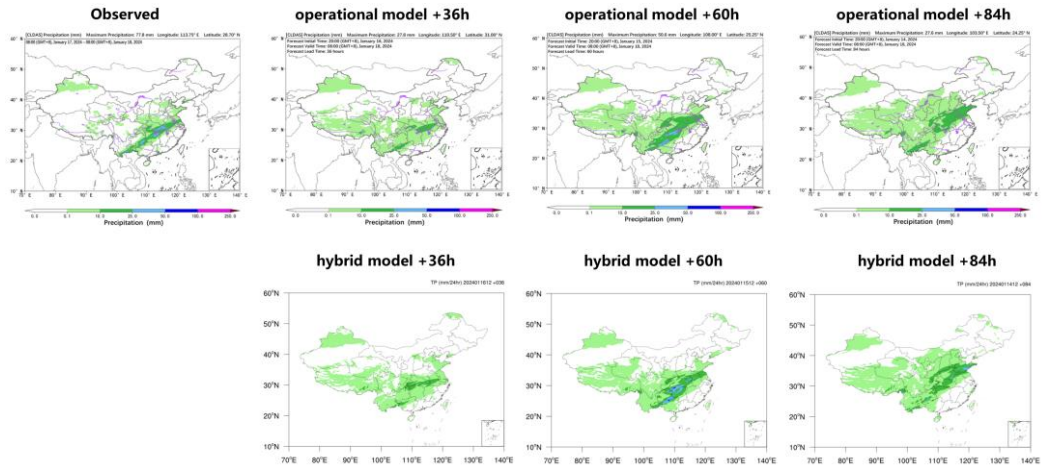

Figure 8. 24-hour accumulated precipitation from 08:00 on January 17 to 08:00 on January 18, 2024.

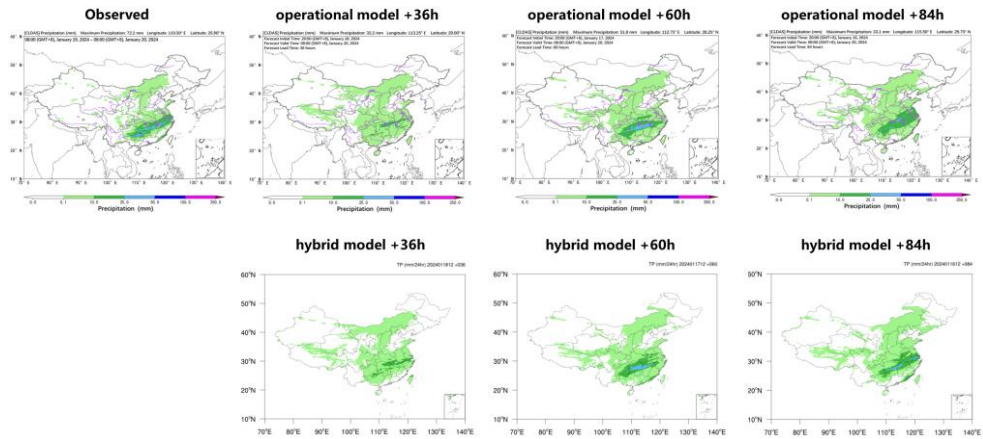

Figure 9. 24-hour accumulated precipitation from 08:00 on January 19 to 08:00 on January 20, 2024.

Verification results for 2-meter temperature forecasts are presented in Figure 10. In January, the absolute error of the hybrid model's 2-meter temperature forecasts was generally comparable to that of the operational model. In July, the hybrid model outperformed the operational model within the first 1.5 days of lead time, but the operational model became more accurate after the second day. Verification results for 10-meter wind speed forecasts are shown in Figure 11. In January, the hybrid model exhibited higher absolute error in 10-meter wind speed forecasts than the operational model. Additionally, its relative error for zonal wind was significantly higher within a 3-day lead time. In July, the hybrid model's absolute error in 10-meter wind speed forecasts also exceeded that of the operational model, with notably higher relative error for meridional wind observed across all validity periods.

A severe cold wave event occurred from January 19 to 20, 2024. For the operational model, its forecast of the temperature change area was consistent with observations but slightly weaker in intensity, particularly failing to capture the temperature change center intensity over central Inner Mongolia. For the hybrid model, the 24-hour forecast was largely consistent with the operational model. However, its 48-hour and 72-hour forecasts showed a slightly stronger intensity of the temperature change center in central Inner Mongolia than the operational model, as illustrated in Figure 12. Verification results for a heat wave event are presented in Figure 13. For the heat wave from July 22 to 23, 2024, the operational model's forecasts of both heat wave range and intensity were weaker than observations. In contrast, the hybrid model's 24-hour forecast of the heat wave range was closer to observations, while its 48–72 hour forecasts indicated a smaller heat wave range in the Xinjiang region compared to the operational model's results.

Typhoon Kaemi formed over the ocean east of the Philippines on the afternoon of July 20, 2024. It then gradually developed and intensified, becoming the first super typhoon of 2024 with its peak intensity reached around the afternoon of July 22. Typhoon Kaemi was characterized by a massive cloud system, strong intensity, and wide-ranging impacts. A comparison of forecast results for Typhoon Kaemi's precipitation is presented in Figure 14(a). Both the hybrid model and the operational model underestimated the intensity of typhoon-induced heavy precipitation. Relative to the operational model, the hybrid model showed no significant improvement in precipitation forecasting, though its predicted precipitation intensity was slightly stronger. Additionally, the hybrid model and the operational model exhibited comparable performance in forecasting Typhoon Kaemi's track and intensity, as illustrated in Figure 14(b).

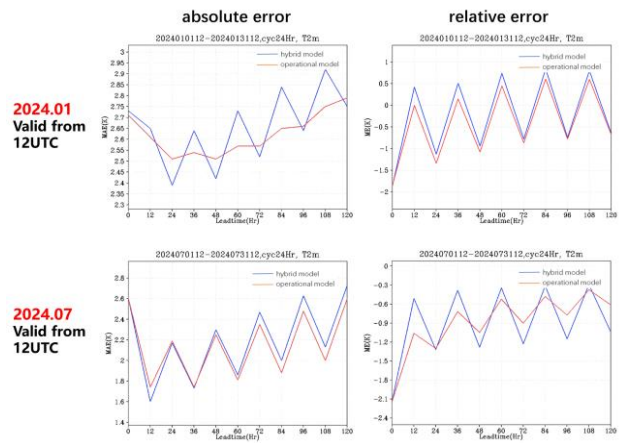

Figure 10. Comparison of 2-meter temperature forecasts: hybrid model vs. operational model.

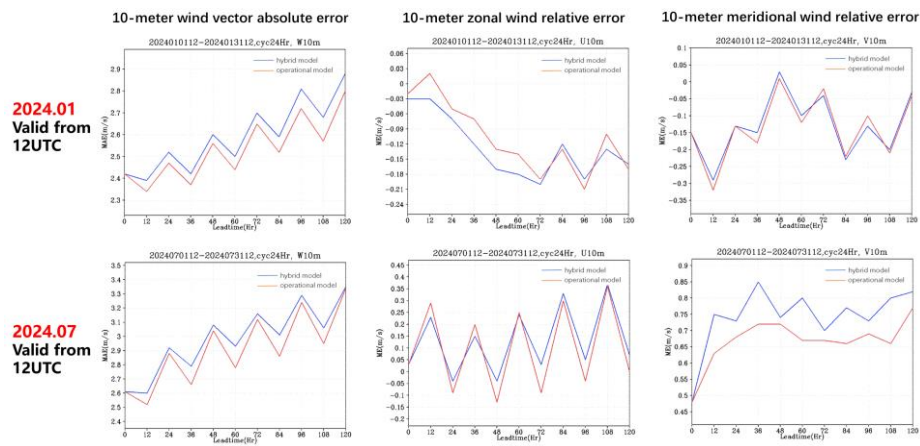

Figure 11. Comparison of 10-meter wind speed forecasts: hybrid model vs. operational model.

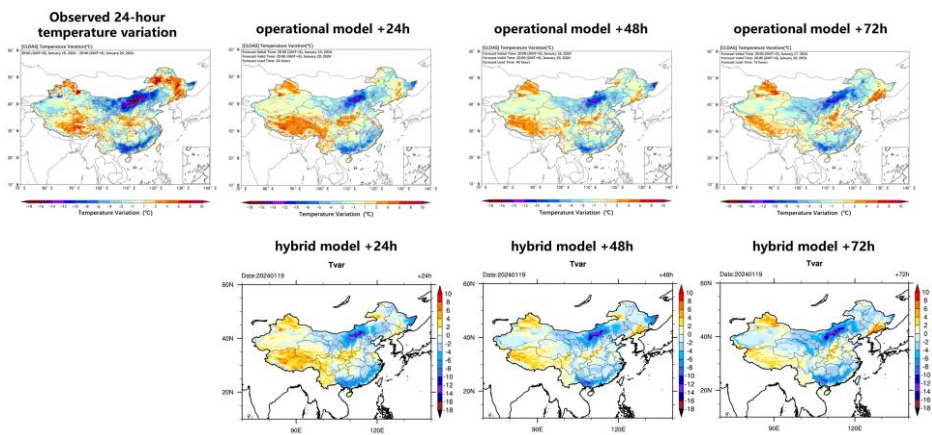

Figure 12. Verification of cold wave forecast performance (January 2024).

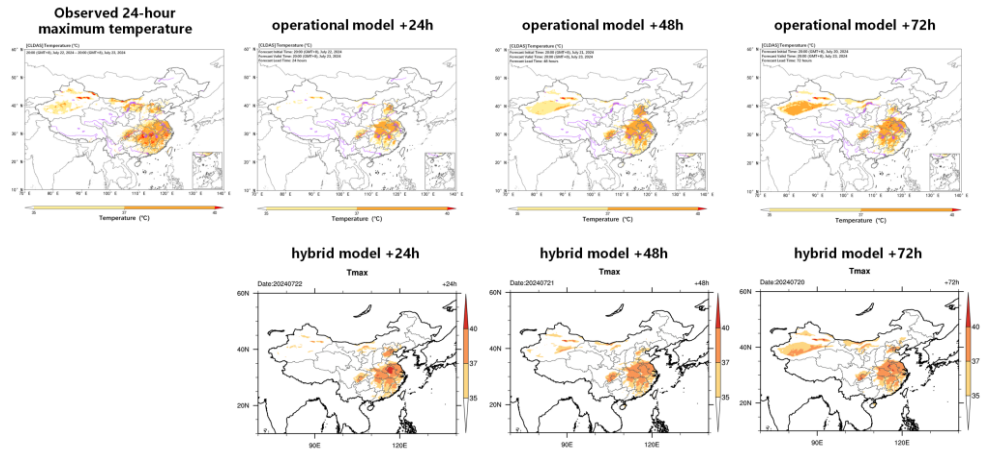

Figure 13. Verification of heat wave forecast performance (July 2024).

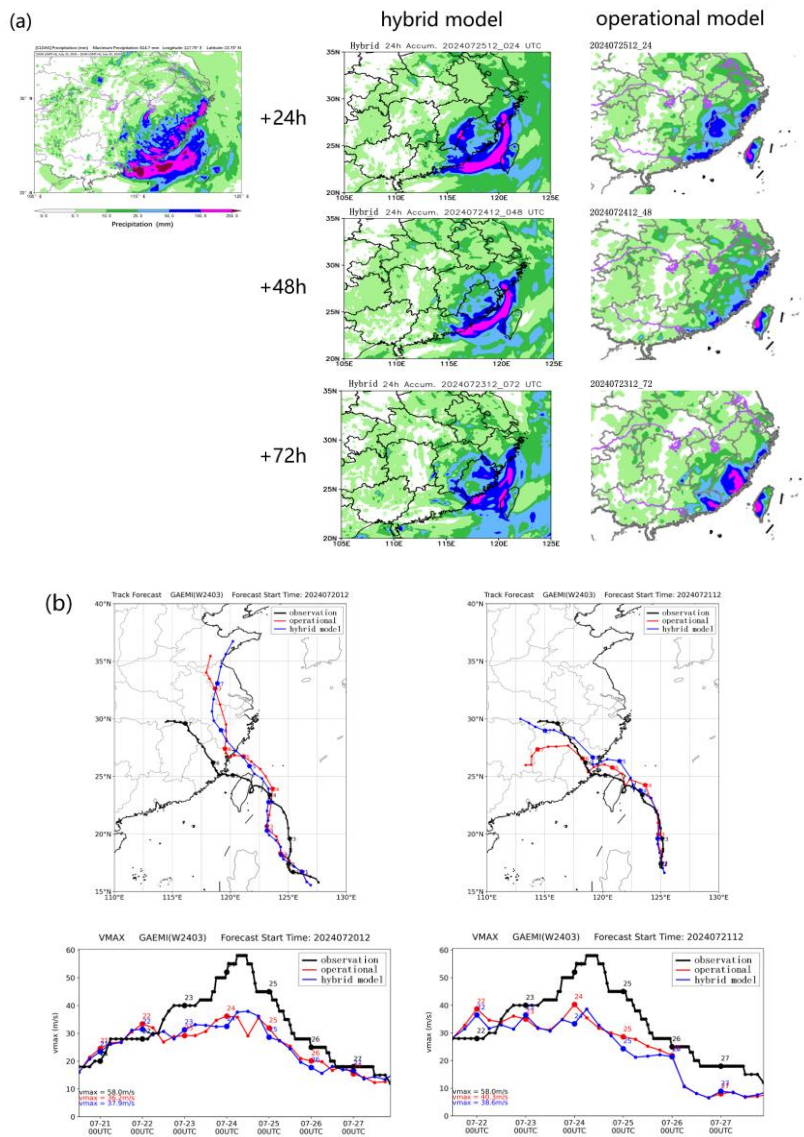

Figure 14. A comparison of Typhoon Kaemi forecasts: (a) precipitation, (b) track and intensity.

Overall, the current hybrid model achieves forecast performance comparable to that of the operational model. However, its performance in forecasting synoptic fields and 10-meter wind speed is slightly diminished — this problem can be addressed by incorporating additional information, such as satellite observation data or LBLRTM physical model. Given that the current deep learning radiation model was initially developed solely to approximate the original RRTMG scheme, the operational reforecast experiments conducted above demonstrate that the predefined objectives have been achieved.

**8. Discussion and future work**

This study developed a machine learning-based parameterization scheme for the radiative physical processes in the CMA-GFS model. It also proposed exclusive solutions to the problems encountered by the hybrid model under operational configurations. The ML emulator was converted to C++ format using Libtorch to enable subsequent invocation by the host model. Stability issues of the hybrid model were resolved via incorporating physical significance, dataset augmentation and experience replay, facilitating stable online coupling. Additionally, this study verified the transferability of the ML emulator across different model resolutions. Even though the ML emulator performed inference on CPU nodes, its computational speed was eight times faster than that of the original physical scheme. A two-month operational reforecast experiment was conducted on the improved hybrid model, followed by a comprehensive evaluation. Results demonstrated that the hybrid model achieved simulation accuracy comparable to the operational model. The hybrid forecasting model built on the CMA-GFS framework, embedding three artificial intelligence modules (one of which is the machine learning radiation module), is now running in real time on a daily basis. A limitation of this approach is that the neural network requires retraining whenever configurational adjustments are made, such as changes to vertical resolution. Future work will focus on further optimizing and refining the neural network model. Using a line-by-line model fully grounded in physical laws to correct the output variables of the training dataset, or integrating observation data. Both strategies aim to improve forecast accuracy.

**Acknowledgment**
This work was supported by the National Natural Science Foundation of China (Grant No. U2242210 and No. U2342220). We also thank Dr. Sun Siyuan and Dr. Zhao Bin for their assistance in preparing the figures.